\documentclass[conference]{IEEEtran}

\usepackage[T1]{fontenc}
\usepackage[utf8]{inputenc}

\usepackage{microtype}
\usepackage{graphicx}
\usepackage{amsmath}
\usepackage{amssymb}

\usepackage{subcaption}
\usepackage{multirow}
\usepackage{makecell}
\usepackage{siunitx}
\usepackage{fontawesome5}

\newcommand{\cmark}{\faCheck}
\newcommand{\xmark}{\faTimes}

\usepackage{booktabs}

\usepackage{soul}
\usepackage{xcolor}
\usepackage{balance}
\usepackage{array}
\usepackage[numbers,sort&compress]{natbib}
\usepackage{url}
\usepackage{xurl}

\usepackage[subtle]{savetrees}

\usepackage{enumitem}
\setlist[itemize]{noitemsep, topsep=0pt, leftmargin=*}
\setlist[enumerate]{noitemsep, topsep=0pt, leftmargin=*}

\newcolumntype{L}[1]{>{\raggedright\arraybackslash}p{#1}}
\newcolumntype{N}{>{\raggedright\arraybackslash}p{1.3cm}}
\newcolumntype{R}[1]{>{\raggedleft\arraybackslash}p{#1}}

\newcommand{\pmark}{\ensuremath{\lozenge}}

\newcommand{\MC}[1]{\shortstack[l]{#1}}

\providecommand{\Description}[1]{}

\begin{document}

\title{PumpSense: Real-Time Detection and Target Extraction of Crypto Pump-and-Dumps on Telegram}

\author{
\IEEEauthorblockN{Ahmed Mahrous, Roberto Di Pietro}
\IEEEauthorblockA{King Abdullah University of Science and Technology (KAUST)\\
Thuwal, Saudi Arabia\\
\{ahmed.mahrous, roberto.dipietro\}@kaust.edu.sa}
}

\maketitle

\begin{abstract}
Cryptocurrency pump-and-dump schemes coordinated via Telegram threaten market integrity. However, existing research addressing this specific threat has not yet produced solutions that combine reliable results with fast response. This is in part due to the absence of publicly available, message-level labeled data, as well as design choices. \\*
In this paper, we address both issues. In particular, we introduce a corpus of over 280,000 Telegram posts from 39 pump-organizing groups, all manually reviewed to identify 2,246 pump announcements and their targeted cryptocurrency and exchange. Leveraging this dataset, we define two tasks: real-time pump-announcement detection and target cryptocurrency/exchange extraction. For detection, we compare two machine-learning models: a lightweight tree-based LightGBM classifier ($F_1$=0.79, latency=9.4 µs/sample) and a transformer-based BGE-M3 ($F_1$=0.83, latency=50 ms/sample). With our proposed approach, we show that message analysis can achieve near-instant pump detection at the level of individual Telegram message windows. Unlike prior work that relies purely on market data and typically detects pumps tens of seconds after abnormal trading activity is observed, our method operates directly on the coordination messages themselves and can be evaluated in microseconds per window on commodity hardware. To our knowledge, we also establish the first benchmark for manipulated coin and exchange extraction. We demonstrate that traditional rule-based extraction methods, widely relied upon in prior literature, are ineffective due to ticker ambiguity. In contrast, LLMs achieve the highest accuracy with a score of 0.91. 

\end{abstract}

\begin{IEEEkeywords}
cryptocurrency pump-and-dump, market manipulation, social media, Telegram, natural language processing
\end{IEEEkeywords}

\section{Introduction}
\label{sec:introduction}

Pump-and-dump schemes, which carry severe legal penalties in traditional financial markets, have increasingly become a structural aspect of cryptocurrency markets. The prevalence and variety of these activities have made cryptocurrency market manipulation a significant field of academic inquiry \cite{EigelshovenUP21}. Cryptocurrency markets have attracted substantial interest from retail investors, who have developed a more speculative trading culture. This, along with the relatively unregulated nature and the ease of issuing low-capital altcoins, creates conditions attractive for manipulators. One common manipulation technique is the \emph{scheduled pump-and-dump}, where organizers announce in advance the time of the event (but not the cryptocurrency). At that scheduled time, they reveal the target cryptocurrency, and participants collectively race to buy it before the organizers sell their holdings. Participants buy, hoping to enter faster than others and capture the price spike.

Telegram, a prominent social-media messaging platform, plays a central role in the coordination of these fraudulent schemes \cite{NizzoliTACTF20}. In numerous public and private groups, anonymous organizers initiate such synchronized buying campaigns targeting cryptocurrencies. Telegram's large, pseudonymous group chats---frequently comprising tens or hundreds of thousands of members---enable high-reach real-time dissemination of pump instructions while minimizing the risk of regulatory detection. Empirical studies have documented thousands of pump events on Telegram between 2018 and 2022, predominantly targeting obscure digital assets with low liquidity \cite{LaMorgia2023Doge, liu2025meme}. These schemes involve large amounts of money---in the order of billions of USD and impact a large portion of cryptocurrencies (e.g., 1 in 4 cryptocurrencies listed on Binance have been previously targeted \cite{charfeddine2024drives}). The unstructured, high-volume nature of Telegram cryptocurrency-related messages---often characterized by specific language---poses challenges for surveillance using natural language processing techniques. 

Despite increased awareness of these fraudulent activities, there remains no publicly available dataset that labels pump announcements at the message level. Prior studies typically record only the timing and names of pumped tokens, without isolating the exact messages that trigger the events. This limitation constrains the development of fraud detection systems, which are necessary for market surveillance or regulatory compliance by social-media platforms, exchanges, regulators, or investors concerned about fraud.

\noindent\textbf{Contributions.}
This paper makes the following contributions to the study of social-media–driven cryptocurrency market manipulation:

\begin{enumerate}
  \item \emph{Message-level pump announcement dataset.}
  We introduce the first publicly available Telegram dataset annotated at the \emph{message-window level} for pump-and-dump coordination. It contains 283{,}017 messages from 39 groups, including 2{,}246 pump-start announcements labeled with target cryptocurrency and exchange. Unlike prior datasets that provide only event timestamps or assets, our corpus captures the exact coordination messages, enabling supervised detection and extraction (Cohen’s $\kappa=0.96$).

  \item \emph{Detection benchmarks.}
  We provide latency-aware benchmarks for message-based pump detection. LightGBM and BGE-M3 achieve up to $F_1=0.83$, with inference times from microseconds (CPU) to milliseconds (GPU), demonstrating feasibility for real-time deployment.

  \item \emph{Extraction benchmarks.}
  We establish the first benchmark for extracting the target cryptocurrency and exchange from pump announcements. Comparing rule-based methods, a long-context transformer, and LLMs, we find that LLMs achieve up to 0.91 joint accuracy and substantially outperform traditional approaches, while remaining practical when applied only to detected windows. We also show that dictionary-based heuristics used in prior work are unreliable.
\end{enumerate}

\noindent \textbf{Roadmap.} In Section~\ref{sec:related_work} we review prior work; Section~\ref{sec:methodology} describes our dataset and methods; Section~\ref{sec:results} reports detection and extraction results; in Section~\ref{sec:ethical_compliance} we cover ethical compliance in data collection and use; Section~\ref{sec:future_limitations} discusses limitations and future directions; and Section~\ref{sec:conclusion} concludes.

\section{Background and Related Work}
\label{sec:related_work}

\subsection{Background}

Cryptocurrency pump-and-dump schemes coordinated via Telegram are well-documented and economically significant. Empirical studies report hundreds to thousands of pump events, targeting a large portion of exchange-listed cryptocurrencies, with price spikes exceeding 60\% within minutes and extreme short-term increases above 1,800\% \cite{xu2019anatomy, hamrick2021examination, charfeddine2024drives}. Profits are highly concentrated among organizers and early participants, while most retail traders incur losses \cite{dhawan2023new}. Long-term effects are negative, with prices declining by roughly 30\% one year after a pump \cite{CloughE23}. 

Prior pump-and-dump detection methods fall into three broad categories. 
\emph{Market-based} approaches rely exclusively on price-, volume-, or transaction anomalies and detect pumps only after abnormal trading activity begins, with reported latencies ranging from tens of seconds to minutes \cite{mahrous2025tstr, bello2023lld, LaMorgia2023Doge, wu11106506}.
\emph{Hybrid} models combine social-media signals with market data but focus on analyzing whether pump-and-dump activity is correlated with social media activity, without attempting to detect pumps through social-media announcements in real-time \cite{mirtaheri2019identifying, xu2019anatomy, nghiem2021detecting, nam2025detecting}.  
\emph{Message-level} approaches analyze social-media text directly, but existing work relies on naive keyword heuristics with limited accuracy and uses proprietary datasets that prevent reproducibility \cite{hu2022sequence, bolz2024machine}. Notably, no prior work reports accuracy for extracting target cryptocurrency and exchange entities at the message level.

\subsection{Prior Approaches}

\begin{table*}[t]
  \small
  \renewcommand{\arraystretch}{1.2}
  \setlength{\tabcolsep}{3pt}
  \caption{Comparison of related work (\cmark=Yes; \xmark=No; \pmark=Partial)}
  \label{tab:pnd-comparison}
  \begin{tabular}{
    L{2.9cm}  
    c c c     
    r r       
    c c       
    L{3.5cm}  
  }
    \toprule
    \textbf{Paper}
      & \MC{\textbf{Manually}\\\textbf{labeled}}
      & \MC{\textbf{Public}\\\textbf{data}}
      & \MC{\textbf{Message-level}\\\textbf{analysis}}
      & \MC{\textbf{\# Messages}}
      & \MC{\textbf{\# Pumps}}
      & \MC{\textbf{Detection}\\\textbf{Model}}
      & \MC{\textbf{Extraction}\\\textbf{Model}}
      & \textbf{Models} \\
    \midrule
    \citet{bolz2024machine}
      & \xmark
      & \xmark
      & \cmark
      &  91\,295
      &   2\,079
      & \cmark\textsuperscript{1}
      & \xmark\textsuperscript{2}
      & Transformer, LLM \\
    \citet{hu2022sequence}
      & \pmark~(5\,050)
      & \cmark
      & \xmark
      &   4\,674\,822
      &     709
      & \cmark
      & \xmark\textsuperscript{5}
      & Tree, Logit, Neural Net \\
    \citet{mirtaheri2019identifying}
      & \pmark~(1\,557)
      & \cmark
      & \cmark
      &   195\,576
      &  62\,850
      & \cmark\textsuperscript{3}
      & \xmark\textsuperscript{5}
      & Support Vector Machine \\
    \citet{la2020pump}
      & \xmark
      & \cmark
      & \xmark
      &        —
      &     900
      & \cmark
      & \xmark\textsuperscript{5}
      & Tree \\
    \citet{CloughE23}
      & \xmark
      & \cmark
      & \xmark
      &  10\,687
      &  10\,687
      & \xmark\textsuperscript{4}
      & \xmark\textsuperscript{5}
      & Statistical analysis \\
    \textbf{This paper}
      & \cmark
      & \cmark
      & \cmark
      &   283\,017
      &   2\,246
      & \cmark
      & \cmark
      & Tree, Transformer, LLM \\
    \bottomrule
  \end{tabular}

\vspace{2pt}
\noindent\begin{minipage}{\textwidth}
  \raggedright \footnotesize
  \textsuperscript{1} Labels messages into 6 classes;\quad
  \textsuperscript{2} No extraction accuracy reported;\quad
  \textsuperscript{3} Code not released;\quad
  \textsuperscript{4} Regex-based detection.
  \textsuperscript{5} Rule-based extraction using dictionaries and regular expressions.
\end{minipage}
\end{table*}

\paragraph{Gaps in Prior Work}
Table~\ref{tab:pnd-comparison} provides a systematic comparison of prior work most relevant to this study---focusing on Telegram-based pump-and-dump datasets, detection, and extraction. This table allows us to highlight the gaps that our work aims to fill. First, although \citet{bolz2024machine} achieve high detection accuracy, their dataset or model is not publicly released to allow for verification of these results. The paper by \citet{bolz2024machine} is the one other than ours that applies LLMs to extract the target cryptocurrency/exchange; however, they do not report the accuracy of their extraction LLM. Their paper is focused on predicting target cryptocurrencies and not post hoc detection. \citet{hu2022sequence} works with a significantly larger Telegram message corpus, but their public dataset does not release the messages they worked with; they only release the timing of pump events and targeted cryptocurrency/exchange. Moreover, their extraction method is based on simple parsing of the text to match known cryptocurrencies and exchanges, a method we show to be highly inaccurate for cryptocurrency extraction; even though they manually label a small sample of announcements, they do not manually verify the target cryptocurrencies. Similarly, \citet{la2020pump} do not release the labeled Telegram messages, nor do they demonstrate the accuracy of their automated extraction method. The detection models these two papers develop utilize market time-series data (e.g., price and volume), and not Telegram messages like our machine-learning models do \cite{hu2022sequence, la2020pump}. \citet{CloughE23} present one of the largest publicly available corpora of pump-and-dump events. While they make their event list public, they do not release the underlying Telegram messages. Moreover, their approach identifies events via simple keyword and regex heuristics without more advanced model testing or manual verification. This raises doubts regarding the accuracy of their event list, particularly since we show that such heuristics can achieve very weak results for target cryptocurrency extraction. \citet{mirtaheri2019identifying} manually labeled only 1,557 messages---likely containing just tens of true pump announcements given the proportion we observe. This makes their data far too small to robustly train or test a detection model. They also do not publish their code, preventing replication or verification. Like most prior work (aside from \cite{bolz2024machine}), they extract both target cryptocurrency and exchange via dictionary/regex lookups; we show in Section \ref{subsec:extraction} that this approach is highly inaccurate. Finally, they report that roughly 30\% of their $\approx$195 000 Telegram messages are pump announcements---an order‐of‐magnitude higher rate than both our measured $\approx$1\% prevalence and the findings of all other studies---calling into question the precision of their detection model. Moreover, Ardia \emph{et al.} publish 1,161 events with success labels but omit chat content \cite{ardia2024twitter}, and Pugoev's master thesis lists 77 pump events with reported impact yet omits original messages \cite{Pugoev2024MasterThesis}. General Telegram datasets such as Pushshift (317 million messages \cite{baumgartner2020pushshift}) and TGDataset (400 million messages across 120,979 channels \cite{la2025tgdataset}) provide large-scale corpora but without pump-specific labels.

\section{Data and Methodologies}
\label{sec:methodology}
This section describes our data collection, annotation process, and methodologies for detecting cryptocurrency pump-and-dump announcements and extracting targeted coins and exchanges.

\subsection{Data Collection and Annotation}
\label{subsec:data}

We collected Telegram posts from 39 public groups known to coordinate cryptocurrency pump-and-dump events. Groups were chosen after a survey of online directories\footnote{e.g., \url{https://www.pump-groups.com/crypto-pump-and-dump}} and academic sources (e.g., \cite{la2020pump}). For each group, we collected all messages, including text content and timestamps. In total, we obtained more than 280,000 messages. 

\begin{figure}[htbp]
  \centering
  \Description{Left panel: a Telegram chat screenshot where a moderator announces a pump on Poloniex with countdown messages. Right panel: a time‑series line chart of the target cryptocurrency's price on Poloniex showing a sharp spike at the pump announcement time.}
  \includegraphics[width=\columnwidth]{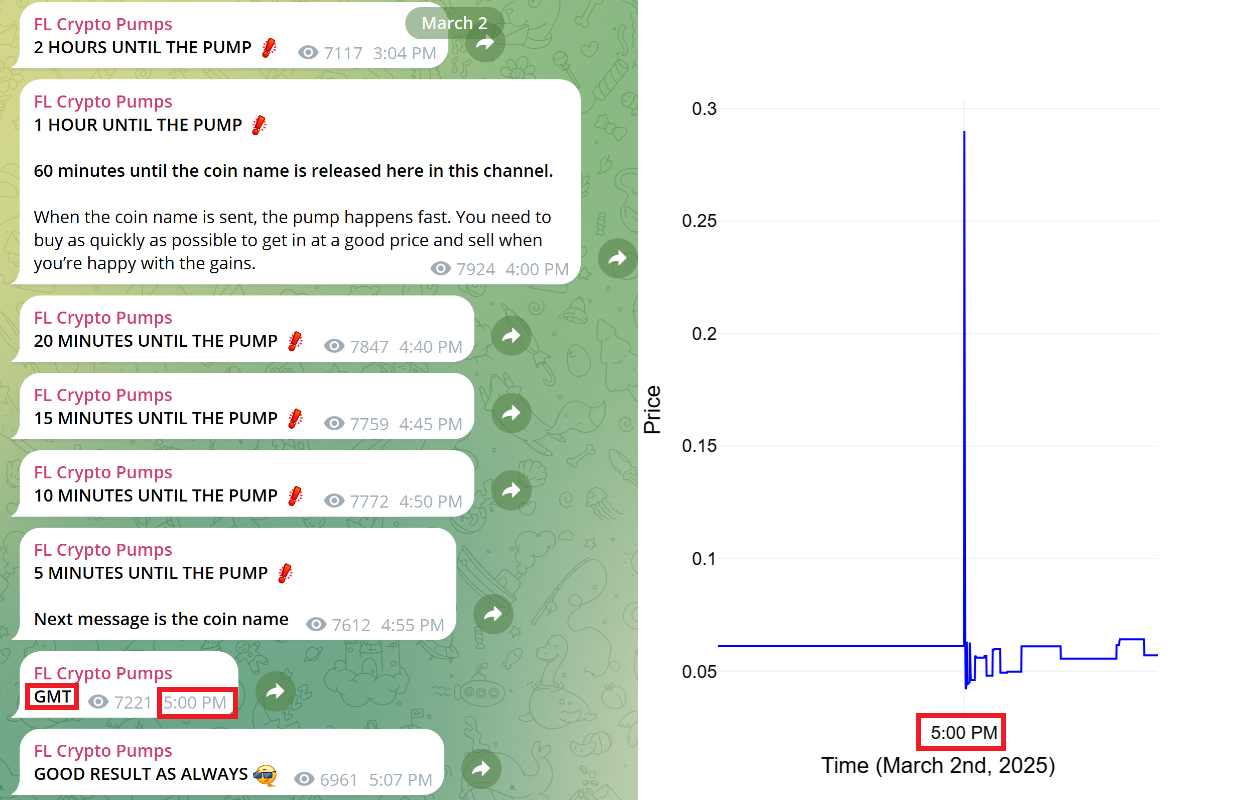}
  \caption{Left: Telegram group announcements regarding a pump-and-dump scheme targeting the GMT coin on Poloniex exchange. Right: the impact of the announcements on the price of the manipulated GMT coin on the Poloniex exchange.}
  \label{fig:pumpdump}
\end{figure}

All messages were manually reviewed to identify pump announcements. A pump announcement was defined as the message that explicitly announced the beginning of a pump. Note that prior to this announcement, there are usually countdowns. After the announcement, moderators usually comment on the status of the pump. Messages deemed to be pump announcements were labeled (1 if the message initiates a pump event, 0 otherwise). The targeted cryptocurrency and exchange were also manually extracted. Annotation was done so that when a pump announcement did not contain either the cryptocurrency or exchange, surrounding messages were inspected for missing information. After labeling, 2,246 messages were marked as pump announcements. Fig~\ref{fig:pumpdump} shows an example of the type of pump announcements we focus on in this paper, along with the sharp price movements it caused. To assess the reliability of manual annotation despite the strong class imbalance (only \(\approx1\%\) of windows are pumps), we drew a stratified sample of 5{,}000 windows (50\% pump, 50\% non-pump) and had a second annotator relabel them. On this balanced set, we observed Cohen's \(\kappa = 0.96\), indicating almost-perfect agreement across both classes. With this high consistency, we deemed the first annotator's labels on the full corpus reliable. Annotation was performed by two annotators with graduate-level training in computer science and prior familiarity with cryptocurrency markets. 

\subsection{Methodologies}
\label{subsec:methodologies}
\subsubsection{Problem Formulation}
The present study addresses two NLP tasks related to pump-and-dump fraud surveillance:
\begin{itemize}
  \item \emph{Pump-announcement classification}: given a fixed-length message window, decide whether that window contains the \emph{first explicit pump start announcement} (the point at which the pumped cryptocurrency is revealed and participants are urged to buy).
  \item \emph{Target extraction}: for windows classified as pump announcements, extract the pumped \emph{cryptocurrency ticker} and \emph{exchange}.
\end{itemize}
Both tasks are evaluated against manually annotated ground-truth labels.

\subsubsection{Data Preparation}
Messages in each Telegram group are ordered chronologically $m_1,\dots,m_n$. 
For every index $i$ we build a window of $11$ messages: the central message $m_i$ plus $5$ earlier and $5$ later  ($m_{i-5},\dots,m_i,\dots,m_{i+5}$). 
At the start and end of the corpus, fewer than five context messages may exist; we keep these shorter windows (no padding). 
All messages in a window are concatenated in time order to form the \emph{window text}.

Each window is labeled $y_i \in \{0,1\}$, where $y_i=1$ if $W_i$ contains the first explicit pump start announcement (per manual annotation), and $y_i=0$ otherwise.

Because adjacent windows overlap heavily (sharing up to 10 messages), a random split of windows into training, validation, and test sets would leak information: the same underlying message could appear in both training and evaluation data.

Instead, we sort windows by the timestamp of their \emph{latest} message and take the earliest $60\%$ for training, the next $20\%$ for validation, and the most recent $20\%$ for testing. This ensures that no message contained in a validation or test window has appeared in the training data and mirrors real-world deployment, where an artificial-intelligence model is trained on past data and then used to detect future pump starts. Windows that cross a split boundary (contain messages on both sides of the split) are assigned to the later split to prevent message-level leakage. The number of such windows is negligible. 

\paragraph{Model Selection Rationale}
Model selection was guided by complementary strengths and by input-length constraints. We include LightGBM for its competitive efficiency in real-time detection, and BGE-M3 and Longformer (LED) for their ability to handle long-context inputs. For extraction, we retain a rule-based method as a prior-work baseline and evaluate LLMs as strong prompt-based extraction models. Long-context support is necessary because windows are built from 11 messages, and approximately 35\% of these windows exceed the standard 512-token limit.

\subsubsection{Pump-announcement Detection}

\paragraph{Feature representation}
Window texts are transformed into TF–IDF vectors over unigrams and bigrams, a common feature representation in natural language processing for text classification.

\paragraph{Baseline Classifier}
We use LightGBM, a tree-based machine-learning model, trained on TF–IDF representations of message windows. Fewer than $1\%$ of messages are positive (\emph{pump}) cases, yielding a highly imbalanced dataset. 
We mitigate this by applying inverse-frequency class weights in the training loss and by evaluating machine-learning models with balanced accuracy, which averages recall across the pump and non-pump classes.

\paragraph{Deep-learning Classifier}
We use BGE-M3, a BERT-based Transformer encoder. Raw message windows are preprocessed to preserve domain-specific tokens while removing unnecessary elements. Preprocessed text is then tokenized using the model's native tokenizer. A linear classification head is then trained on this data. We use a maximum input length of 1024 tokens, which covers over 95\% of input windows.

\subsubsection{Target Information Extraction}
We compare four approaches to extract the target cryptocurrency and exchange from a window $W_i$:

\paragraph{Rule-Based Baseline Extractor}
We prepare a list of 12\,000+ cryptocurrency tickers and 43 exchange names. This baseline extraction method identifies the first occurrence in the window text matching one of the listed cryptocurrency tickers or exchanges.

\paragraph{Deep-learning Extractor}
We adopt a sequence-to-sequence Longformer Encoder-Decoder (LED) machine-learning model for structured extraction, and use \texttt{LED-base-16384}. Each 11-message window is paired with a target string of the form "\texttt{<cryptocurrency>}\textbar{}\texttt{<exchange>}". During fine-tuning, the model is trained to map each concatenated 11-message window directly to its corresponding "\texttt{<cryptocurrency>}\textbar{}\texttt{<exchange>}" string.

\paragraph{DeepSeek LLM}

An LLM-based approach was also implemented using the \texttt{DeepSeek-V3-0324} model through its API\footnote{\url{https://api.deepseek.com}}. The prompt consists of the full window text and instructs the model to extract the cryptocurrency ticker and exchange name in a structured format, where the cryptocurrency is prefixed with \texttt{cryptocurrency:} and the exchange with \texttt{Exchange:}. The model's response is scanned for these markers. Extracted values are normalized to lowercase and compared against the ground-truth labels using the same evaluation criteria as the rule-based model: cryptocurrency, exchange, and joint accuracies are reported accordingly. The prompt instructs the model to extract the cryptocurrency ticker and exchange in a structured format; the exact prompt is provided in the released code.

\nocite{chegenizadeh2025scalable}




\paragraph{Gemini LLM}
We follow exactly the same methodology as for DeepSeek, replacing the model with Google’s Gemini LLM (version \texttt{2.5-pro-preview}).%
\footnote{\url{https://ai.google.dev/gemini-api}}. The prompt and all other details remain identical.

\paragraph{GPT LLM}
We also evaluate the \texttt{GPT-4.1} model via its API\footnote{\url{https://platform.openai.com/docs/models/gpt-4.1}}. We used the same prompt template as for DeepSeek and Gemini and followed the same methodology.

\paragraph{Evaluation}
For each method, we compute cryptocurrency accuracy, exchange accuracy, joint accuracy (both correct), and inference time per sample.

Here, accuracy is defined as the number of correctly predicted instances divided by the total number of pump announcement instances in the test set.
All models are evaluated on the same held‐out 20\% test set of 420 pump windows. 

Inference times are reported to compare the models' efficiencies. All experiments were conducted on a machine equipped with an NVIDIA GeForce RTX 4070 Laptop GPU (8GB VRAM) and 32GB RAM.

\section{Results}
\label{sec:results}
This section presents descriptive statistics of our dataset, followed by results from the pump detection and target coin/exchange extraction tasks.

\subsection{Dataset Stylized Facts}

The dataset contains a total of 283,017 messages collected from 39 different Telegram groups known for coordinating pump-and-dump fraud. Table \ref{tab:dataset_summary} highlights key characteristics of this 283,017-message corpus. The data spans a period of approximately six years, with the first message dated mid-2017 and the last recorded in mid-2023. The average message length is 244 characters, with pump announcement messages being, on average, about 100 characters longer than non-pump messages.

\begin{table}[htbp]
  \centering
  \footnotesize
  \setlength\tabcolsep{1pt}
  \renewcommand{\arraystretch}{1.1}
  \caption{Overview of our publicly released Telegram pump‐and‐dump dataset.}
  \label{tab:dataset_summary}
  \begin{tabular}{
    L{3.2cm}  
    L{1.8cm}  
    L{3.cm}  
  }
    \toprule
    \textbf{Statistic} & \textbf{Value} & \textbf{Notes} \\
    \midrule
    Total messages collected          & 283\,017        & From 39 Telegram groups \\
    Pump announcements                & 2\,246          & Manually labeled \\
    Cancelled pump events             & 140             & Manually labeled \\
    Unique cryptocurrency tickers    & 604             & Manually extracted \\
    Unique exchanges targeted         & 14              & Manually extracted \\
    Image‐based pump announcements    & 255             & $\approx11\%$ include images \\
    Collection period                 & 2017–2023       & $\approx$6 years \\
    Avg.\ message length              & 244 chars       & Std.\ dev.\ $\approx76$ chars \\
    Avg.\ pump msg length             & 344 chars       & $\approx100$ chars longer \\
    Top 10 groups (by \# pumps)       & 70\% of pumps   & Heavy‐tail distribution \\
    Peak hours for announcements      & 15:00–17:00 UTC & 70\% occur then \\
    Peak days for announcements       & 40\%            & Saturdays \& Sundays (Weekends) \\
    \bottomrule
  \end{tabular}
  \renewcommand{\arraystretch}{1.0}
\end{table}

After reviewing all the messages, 2,246 pump announcements---fewer than 1\% of all---were manually identified and labeled. In addition, 140 pump events were labeled as cancelled based on moderator messages. In total, 604 different cryptocurrency tickers and 14 different cryptocurrency exchanges are targeted. For each manually identified pump announcement, the targeted cryptocurrency and exchange were manually extracted. Throughout the dataset, some cryptocurrencies are targeted up to 40 times. 

Temporal patterns reveal that around 70\% of pump announcements occur between 15:00 and 17:00 UTC, and nearly 40\% take place on weekends (Saturdays and Sundays). This finding is in line with previous research \cite{charfeddine2024drives}. The first labeled pump announcement in the dataset occurred in June 2017, and the most recent one in April 2023. 

The distribution of pump announcements across groups is heavily skewed: 10 influential groups are responsible for approximately 70\% of all identified pump events. This heavy-tailed distribution of pumps per group was also documented in previous research \cite{hamrick2021examination}. A total of 255 pump announcements include an attached image. Additionally, we manually write some unstructured notes for some of the pump events (e.g., records of insiders gaining early knowledge of the target cryptocurrency). While efforts were made to identify all pump announcements, some may have been missed due to the large size of the dataset. 

\subsection{Detection}
\label{subsec:detection}

\begin{figure}[b]
  \centering
  \Description{Two confusion matrices side by side. (a) Confusion matrix for LightGBM: true negatives = 51354, false positives = 1394, false negatives = 436, true positives = 3334. (b) Confusion matrix for BGE‑M3: true negatives = 52379, false positives = 369, false negatives = 828, true positives = 2942.}
  \subcaptionbox{Confusion matrix for LightGBM\label{fig:lightgbm}}{%
    \includegraphics[width=0.4\columnwidth]{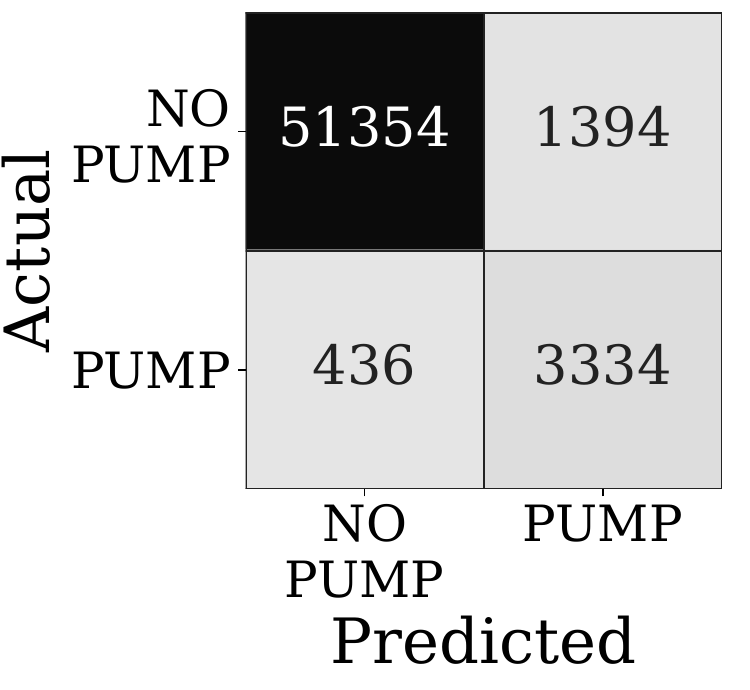}}%
  \hfill
  \subcaptionbox{Confusion matrix for BGE\label{fig:bge}}{%
    \includegraphics[width=0.4\columnwidth]{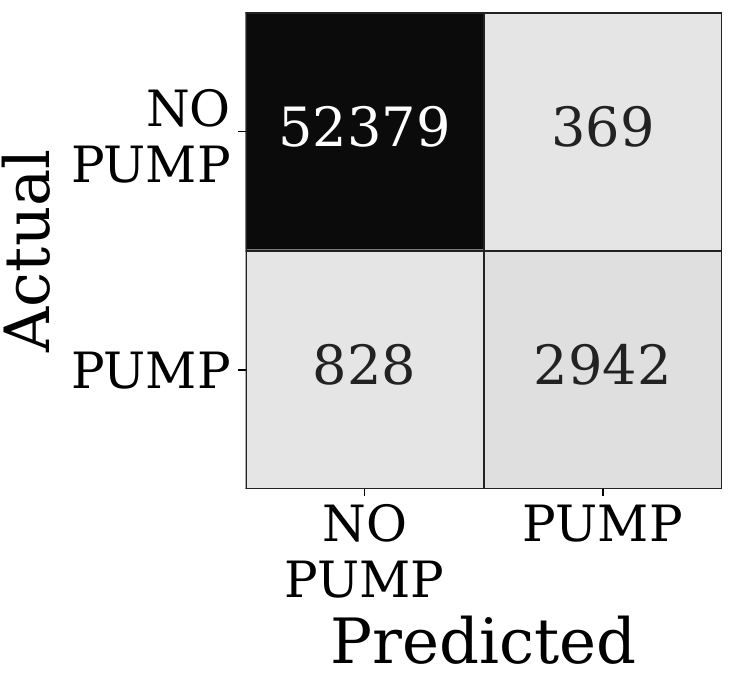}}%
  \caption{Confusion matrices for pump‐start detection.}
  \label{fig:confusion_matrices}
\end{figure}

The performance of two machine-learning models for detecting pump-and-dump announcements is detailed in Table \ref{tab:pump_detection_results}, with corresponding confusion matrices in Figure \ref{fig:confusion_matrices}. Both models reach comparable overall effectiveness: BGE--M3 outperforms LightGBM in $F_1$ by only $0.04$ (0.83 vs.\ 0.79).

\begin{table}[t]
\centering
\caption{Pump‐start classification performance on the test set. Metrics are for the minority "Pump" class. Time values represent inference latency in seconds per sample.}
\label{tab:pump_detection_results}
\sisetup{table-number-alignment = center, 
         scientific-notation = true, 
         retain-explicit-plus = true}
\resizebox{\columnwidth}{!}{%
  \begin{tabular}{lcccc}
    \toprule
    \textbf{Method} & \textbf{F$_1$} & \textbf{Precision} & \textbf{Recall} & \textbf{Time} \\
    \midrule
    LightGBM & 0.79 & 0.71 & 0.88 & \num{9.4e-6}  \\
    BGE-M3   & 0.83 & 0.89 & 0.78 & 0.05  \\
    LLM (GPT)  & 0.65 & 0.68 & 0.62 & 0.6  \\
    \bottomrule
  \end{tabular}%
}
\end{table}

LightGBM is more permissive in this configuration, yielding higher recall ($0.88$) but lower precision ($0.71$), whereas BGE--M3 is more conservative, trading some recall ($0.78$) for substantially higher precision ($0.89$). These operating points are not fixed: rebalancing class weights, adjusting loss costs, or simply shifting the decision threshold can move either model along its precision--recall curve.  

In our setting, missing a true pump start is worse than flagging a false start. False negatives deprive defensive systems of the opportunity to prevent fraud, whereas false positives can be checked by a more precise (more expensive) artificial-intelligence model or a human. Since pump announcements are rare---less than 1\% of all traffic---a recall-biased machine-learning model would generate a manageable amount of false positives.

In terms of speed and efficiency, LightGBM shows a clear advantage. LightGBM classifies a window in a near-instantaneous \SI{9.4}{\micro\second} on a CPU core, supporting cheap real-time coverage of busy channels. Prior work based solely on market-microstructure features (e.g., \citet{LaMorgia2023Doge}) reports detection latencies on the order of tens of seconds \emph{after} abnormal trading activity begins; our approach instead operates directly on the coordination messages, and thus can in principle raise alerts before or at the onset of the price reaction. These two lines of work are complementary: textual detection provides an early social-signal layer, while market-based methods monitor downstream price dynamics. BGE--M3 needs \SI{50}{\milli\second} on a GPU; deploying it on every window would be less practical under heavy traffic. 

Given the similar $F_1$ scores, the low latency and higher recall of LightGBM make it a more practical first-line detector. In this case, a simple tree-based machine-learning model (LightGBM) is therefore preferable to the slower deep-learning model (BGE-M3) for pump-and-dump surveillance on Telegram. In addition to $F_1$ scores, the LightGBM and BGE-M3 detectors achieve ROC–AUC values of approximately 0.91 and 0.92, respectively, indicating stable performance across decision thresholds.

We also evaluate two context constructions: symmetric (\emph{offline}) windows, which include both past and future messages around the target message, and trailing (\emph{online}) windows, which include only past/current messages. Ablations over window size (3/7/11) and TF-IDF dimensionality (10k/15k/20k) show that size 3 underperforms ($F_1 \approx 0.64$--$0.70$), while sizes 7--11 perform best ($F_1 \approx 0.79$--$0.82$), with the top setting at symmetric size 11 and 20k features ($F_1 = 0.820$).

We also experimented with LLMs for pump-start detection, reaching a balanced accuracy of 0.73 in 0.6s/window. However, performance proved volatile---accuracy significantly varied depending on the specific LLM and prompt used. The combined cost, latency, and unpredictability make LLMs impractical as first‑line detectors at scale; instead, we reserve them for the $\approx$1\% of windows flagged by our lightweight machine-learning model. It is worth noting that while newer generations of LLMs will undoubtedly improve in speed and capabilities, the performance gap we observe here is structural. The latency difference between LightGBM (microseconds) and LLMs (seconds or milliseconds) spans several orders of magnitude. Even if future LLMs become significantly faster, they are unlikely to match the throughput of a lightweight tree-based model. Consequently, the architectural choice of using a specialized, supervised model for first-line filtration remains valid regardless of incremental advancements in generative AI.

\paragraph{Cross-Validation Robustness}
To assess robustness, we performed a time-ordered 5-fold cross-validation for LightGBM across three feature counts and three random seeds (45 runs total). Recall remained stable, with mean recall of 0.87 and standard deviation below 0.04 across all 5-fold configurations.

\paragraph{Event-Level Detection Delay}
We also measured event-level detection delay. The detector identifies 88.8\% of pump announcements exactly at the announcement message (0-message delay), and 91.6\% within five messages. This indicates that practical end-to-end alert latency is primarily constrained by inference speed, which we already report.

Table~\ref{tab:string_diffs} contrasts how often selected phrases appear in pump‐start (\texttt{pump\_1}) versus background (\texttt{pump\_0}) .  Many expressions are overwhelmingly more common in pump starts. Phrases indicating future timing, such as "will be" and "minutes left," appear much more frequently in pump-start windows. Direct markers like "cryptocurrency name," "pump," and "exchange" are also common, along with collective action phrases such as "we," "everyone," and "make sure."

\begin{table}[htbp]
\centering
\caption{Most discriminative phrases appearing in pump announcements, ranked by the percentage-point difference between pump-start windows and background message windows. Percentages denote the fraction of messages in which a phrase appears. Difference is computed as Pump minus Background.}
\label{tab:string_diffs}
\resizebox{\columnwidth}{!}{%
\begin{tabular}{lccc}
\toprule
\textbf{Phrase} & \textbf{Pump (\%)} & \textbf{Background (\%)} & \textbf{Difference (\%)} \\
\midrule
will be        & 82.06 & 14.84 & +67.22 \\
exchange       & 69.81 & 12.74 & +57.08 \\
left           & 59.90 & 9.40  & +50.50 \\
coin           & 97.23 & 52.17 & +45.05 \\
pump           & 88.72 & 46.19 & +42.53 \\
minutes left   & 40.53 & 1.97  & +38.56 \\
\bottomrule
\end{tabular}
}
\end{table}

These large differences mean that a bag-of-words or TF--IDF vector could by itself carry a strong signal. A tree ensemble such as LightGBM can learn rules like if "minutes left" occurs more than 3 times and "coin name" occurs at least 1 time, then predict pump. 

Thus, much of the detection task could be reduced to spotting a small vocabulary of discriminative words and phrases---a setting where lightweight artificial-intelligence models can excel. More sophisticated models can still help when announcements are obfuscated or phrased unusually, but the baseline performance of LightGBM demonstrates that costly deep-learning models are unnecessary for most traffic.

\subsection{Extraction}
\label{subsec:extraction}
For the information–extraction task—identifying the pumped cryptocurrency and the target exchange given a pump announcement—the rule-based baseline is inadequate. Although it retrieves around 60\% of exchanges accurately, it fails to extract any target cryptocurrency from the 420 pump announcements in the test set (Table~\ref{tab:extraction_results}).  

\begin{table}[t]
\centering
\caption{Extraction performance on the unseen test set of 420 pump announcements. Accuracy is computed as the fraction of correct predictions. Inference time is reported in seconds per sample.}
\label{tab:extraction_results}
\resizebox{\columnwidth}{!}{%
  \begin{tabular}{lccc}
    \toprule
    \textbf{Method} & \multicolumn{2}{c}{\textbf{Accuracy}} & \textbf{Time (s)} \\
    \cmidrule(lr){2-3}
                    & \textbf{Cryptocurrency} & \textbf{Exchange} &  \\
    \midrule
    Rule-based     & 0.00 & 0.61 & 0.0005 \\
    Longformer     & 0.51 & 0.60 & 0.18 \\
    DeepSeek    & 0.92 & 0.87 & 7.33 \\ 
    Gemini & 0.95 & 0.93 & 14.35 \\ 
    GPT        & 0.96 & 0.92 & 2.18 \\
    \bottomrule
  \end{tabular}%
}
\end{table}

Two factors could explain this gap. First, the candidate set of exchanges is small (43 exchanges), whereas the list of possible cryptocurrency tickers exceeds 12\,000. Second, while the name of exchanges is mostly unambiguous in everyday language, many tickers coincide with common words (e.g., cryptocurrencies with the 
tickers \texttt{FOR}, \texttt{AT}, \texttt{PUMP}), causing the naive pattern-matcher to misinterpret ordinary text as a ticker—e.g., deciding that any window containing the word \emph{for} must target the cryptocurrency with the ticker \texttt{FOR}. Moreover, when multiple tickers occur in the same window, the baseline always returns the first match. This finding highlights a critical reliability issue in prior datasets constructed using similar keyword-based heuristics.

Deep-learning approaches can overcome these limitations. Longformer, a transformer-based deep-learning model with long context support, improves cryptocurrency extraction to 0.51 and exchange extraction to 0.60, with a joint accuracy of 0.37. Though a clear improvement over the baseline, these numbers remain modest---highlighting the limits of medium-sized deep-learning models in handling this entity extraction task. 

LLMs such as DeepSeek, Gemini, and GPT-4.1 show substantial gains in extraction accuracy. DeepSeek achieves 0.92 cryptocurrency accuracy, 0.87 exchange accuracy, and 0.83 joint accuracy; Gemini improves further to 0.95, 0.93, and 0.91, respectively; GPT-4.1 tops cryptocurrency extraction at 0.96 and joint accuracy closely behind at 0.90 with 0.92 exchange accuracy. These LLMs vary in inference cost: DeepSeek requires approximately 7 seconds per announcement, Gemini around 14 seconds, while GPT-4.1 achieves only about 2.18 seconds per announcement. Note, however, that these measured inference times could also reflect differences in APIs, not just the intrinsic computational speed of each LLM. Since extraction is performed only on a small subset of flagged windows (about 1\% of the total message stream), the computational expense for each LLM remains practical in real-world deployment.

\paragraph{LLM Stability}
To assess extraction stability, we repeated GPT-based extraction under the same setup and observed very low variability, with joint-accuracy standard deviation below 0.003.

\subsubsection{Error Analysis}
Despite the strong performance of LLMs, errors still occur. Out of 420 test samples, Gemini correctly extracted both the cryptocurrency and the exchange in 90.7\% of cases. We manually analyzed the 39 failure cases to identify their causes. Extraction mistakes arise from a set of ambiguities. First, the predictions and labels can refer to closely related but not identical entities (e.g., Binance vs. Binance Futures). Second, the manually annotated labels may be written slightly differently than the predictions (e.g., gateio vs. gate.io). These are often genuine ambiguities where both forms are plausible depending on context. Third, several windows contain multiple pumps, and the LLM may extract a different pump than the one (i.e., the first pump) identified by the annotator. Fourth, errors arise when the exchange is not mentioned directly in the analyzed window but is implied from prior messages, pinned posts, or the group name.

Errors also occur when non-pump cryptocurrencies or exchanges are mentioned alongside the true pump target, misleading the LLM. In rare cases, the cryptocurrency ticker is entirely absent from the text (e.g., shown only in an image). Addressing these errors would increase the reported accuracy even further, and is possible through additional preprocessing of the data or by using different prompts.

In summary, while simple heuristics may occasionally extract exchange mentions, they fail to handle ambiguity. Accurate cryptocurrency extraction, in particular, requires deep contextual understanding. LLMs provide this capability and remain feasible for deployment given the limited number of announcements requiring extraction. However, to maximize their effectiveness, attention should be given to the design of preprocessing steps and the formulation of prompts, both of which can influence extraction quality in ambiguous cases.

\section{Ethical Compliance}
\label{sec:ethical_compliance}
Our data collection adheres to Telegram's Terms of Service \cite{TelegramTOS} and the Telegram API Terms of Service \cite{TelegramAPI}, neither of which prohibits collecting messages from public channels. We refrained from any automated actions that could disrupt the platform's activity or go against its terms. Under GDPR Article 85's academic-research exemption \cite{GDPR85}, we process only publicly available information for scholarly purposes, and---since no private user data is involved---our procedures fall under the GDPR's legitimate‑interest basis (Article 6(1)(f)) \cite{GDPR61f}. In line with the GDPR principle of data minimisation, we restrict our analysis to administrator-generated posts and discard any material not essential to our research objectives. Consequently, no further approval was necessary as per our institutional review board's policy.

\section{Practical Deployment and Surveillance Pipeline} 
\label{sec:pipeline}

Our methods are designed for real-time deployment to monitor and mitigate cryptocurrency pump-and-dump schemes coordinated on Telegram. This section describes how the proposed detection and extraction models can be integrated into a scalable surveillance pipeline, and how such a system can be used both operationally and to continuously expand the dataset.
Each operational stage in this pipeline is supported by the evaluations in Section~\ref{sec:results}: first-line detection is backed by Section~\ref{subsec:detection} and Table~\ref{tab:pump_detection_results}, and target extraction is backed by Section~\ref{subsec:extraction} and Table~\ref{tab:extraction_results}.

\subsection{Discovery of Pump-Organizing Groups}
Identifying public Telegram groups that coordinate pump-and-dump schemes is a practical challenge, as new groups emerge over time. In practice, group discovery can be bootstrapped from a small seed set of known pump groups, such as those used in this study. Additional candidates can be identified using network-based signals, including forwarded messages and invite links. Lightweight text classifiers trained to recognize pump-related language can then be used to filter candidate groups. Groups that exhibit sustained pump-like behavior can be added to the monitoring set. 

\subsection{Continuous Monitoring and First-Line Detection}
Once candidate groups are identified, all messages are processed in chronological order. Messages are segmented into overlapping sliding windows and passed to a lightweight first-line detector, such as the LightGBM model evaluated in this work. Due to its microsecond-level inference time, this detector can scale to large numbers of groups and messages at low computational cost. In practice, only a small fraction of windows (approximately 1\% in our dataset) are flagged as potential pump-start announcements.

\subsection{Confirmation and Information Extraction}
Windows flagged by the first-line detector are further analyzed using more precise but computationally expensive models. In our design, LLMs are applied only to these candidate windows to extract structured information about the target cryptocurrency and exchange. This cascading design limits LLM usage to a very small subset of messages, making high-accuracy extraction practical in real-time settings. Optional downstream checks---such as verifying whether abnormal trading activity follows shortly after the detected announcement---can further increase confidence without introducing significant delay, as explored in prior market-based pump detection studies (e.g., \cite{LaMorgia2023Doge, xu2019anatomy}).

\subsection{Human-in-the-Loop Verification and Dataset Expansion}
The same pipeline can be used to continuously expand the labeled dataset. A sample of newly detected pump events is queued for manual review, where analysts confirm detections and correct extracted entities if needed. Verified events are then added to the corpus and used for periodic retraining of both detection and extraction models. This human-in-the-loop process supports adaptation to evolving language and obfuscation strategies. 

\subsection{Mitigation and Applications}
A deployed system based on this pipeline supports several mitigation strategies. Exchanges can use alerts to monitor suspicious activity, investors can be warned about manipulation attempts, and regulators or platform operators can use records of coordination behavior for further investigation. By focusing on explicit coordination messages rather than price prediction, the system provides early evidence of manipulation intent and complements market-based surveillance tools that monitor market data. 

Overall, this deployment-oriented design shows that the proposed methods provide a practical and scalable foundation for real-time detection, analysis, and mitigation of social-media-driven market manipulation.

\section{Limitations and Future Work}
\label{sec:future_limitations}

This study focuses on a specific manipulation pattern, namely pre-announced pump-and-dump schemes. Other manipulation strategies may exhibit different coordination dynamics and could require adapted detection and extraction methods. Our dataset consists of Telegram messages from 39 public pump-oriented groups. While this scope may appear limited, prior work shows that pump activity is highly concentrated, with a small number of influential groups responsible for most events \cite{charfeddine2024drives, hamrick2021examination}. Moreover, analyzing many groups introduces substantial linguistic diversity. We focus on public groups because they enable coordination on a larger scale and are observable by exchanges and regulators. It is often the case that public groups repost content from private groups with short delays. As a result, models trained on public data are likely to transfer to private groups. The dataset spans mid-2017 to mid-2023, covering multiple market regimes and evolving language use; the endpoint reflects practical constraints related to data access and manual annotation. 

We also experimented with LLMs for pump-start detection, reaching a balanced accuracy of 0.73 in 0.6s/window. However, performance proved volatile---accuracy significantly varied depending on the specific LLM and prompt used. The combined cost, latency, and unpredictability make LLMs impractical as first‑line detectors at scale; instead, we reserve them for the $\approx$1\% of windows flagged by our lightweight machine-learning model. 

Finally, we evaluate strong, widely used LLMs available at the time of experimentation. Because LLM capabilities evolve rapidly, we frame our conclusions as comparative benchmark results rather than permanent state-of-the-art claims. Given the rapid pace of development in the field, we anticipate that subsequent model generations will offer superior capabilities at lower inference costs. We argue that such advancements will not invalidate our findings but rather reinforce them: as LLMs become cheaper and more effective, the viability of the proposed extraction pipeline will only increase, further widening the performance gap between LLM-based extraction and traditional rule-based baselines.

Several directions for future work follow naturally. About 10\% of pump announcements include images, suggesting opportunities for multimodal detection. The dataset also contains 140 manually annotated cancelled pump events, which could be analyzed to better understand coordination failures and strategic behavior. Beyond Telegram, future studies could examine cross-platform manipulation on Discord or Twitter, where recent enforcement actions targeted similar schemes \cite{justice2022eight}. Adversarial adaptation is another important direction: studying how manipulators attempt to evade detection could inform more robust, continuously learning systems. While our LightGBM classifier achieves strong performance ($F_1$=0.79), false positives (29\%) and missed pumps (12\%) remain. Future work could analyze these errors and utilize hybrid pipelines that apply more accurate but slower models only to ambiguous cases. 

\section{Conclusion}
\label{sec:conclusion}
Market manipulation in cryptocurrency markets remains a pressing issue, including pump-and-dump schemes organized on Telegram. Research shows these operations involve billions of dollars, affect millions of investors, and target up to one in four cryptocurrencies on major exchanges such as Binance. In this paper, we tackle two key problems: (i) detecting pump-and-dump announcements and (ii) identifying the cryptocurrency and exchange that the manipulation is targeting. We publicly release a first-of-its-kind corpus of Telegram messages manually annotated for pump announcements, tickers, and exchanges, with manual labels achieving almost‑perfect inter-annotator agreement (Cohen's $\kappa = 0.96$), underscoring the dataset's reliability.

For detection, we implement two machine-learning models---LightGBM ($F_1$=0.79) and BGE‑M3 ($F_1$=0.83)---demonstrating that pump announcements can be identified with high accuracy at near-instant speeds, making large-scale, real-time monitoring of Telegram pump groups feasible. Whereas prior pump-and-dump detectors based purely on market data report latencies on the order of tens of seconds after abnormal trading activity begins, our text-based detector operates directly on the coordination messages and can be evaluated in microseconds per window. Combining such early social-signal detection with downstream market-based detectors is a promising direction for large-scale, multi-layered surveillance.  For target identification, we find that LLMs (DeepSeek, Gemini, GPT) substantially outperform a Longformer baseline (accuracy = 0.83–0.91 vs. 0.37), albeit at a higher inference cost. Importantly, since only about 1\% of all messages are detected as pump announcements, the more costly LLM inference is applied to a very small subset---making LLM‑based extraction practical in real‐time systems.

To our knowledge, these experiments provide the first message-based benchmark for Telegram pump-and-dump detection and target coin/exchange identification. We release our dataset and code to support replication and further research. Our work enables real‑time social‑media surveillance systems to flag market-manipulation schemes and protect investors before harm occurs.

\section*{Code and Data Availability}
Code and data are publicly available on GitHub and archived on Zenodo \cite{mahrous2026repro}.

\balance
\bibliographystyle{IEEEtranN}
\bibliography{references}

\end{document}